\documentclass[preprint,12pt]{elsarticle}

\usepackage{amsmath,amsfonts,amssymb}
\usepackage{bm}
\usepackage{bbm}
\usepackage{array}
\usepackage{graphicx}
\usepackage{multirow}
\usepackage{booktabs}
\usepackage{textcomp}
\usepackage{url}
\usepackage{xcolor}
\usepackage{todonotes}
\usepackage[ruled,vlined]{algorithm2e}
\usepackage[hidelinks]{hyperref}

\newcommand{\MLC}{\texttt{MLC}}
\newcommand{\CV}{\texttt{CV}}
\newcommand{\NLP}{\texttt{NLP}}
\newcommand{\BCE}{\texttt{BCE}}
\newcommand{\LSEP}{\texttt{LSEP}}
\newcommand{\ZLPR}{\texttt{ZLPR}}
\newcommand{\MSC}{\texttt{MSC}}
\newcommand{\AAPD}{\texttt{AAPD}}
\newcommand{\PASCAL}{\texttt{PASCAL}}
\newcommand{\COCO}{\texttt{COCO}}
\newcommand{\SGD}{\texttt{SGD}}

\newcommand{\lreg}{\mathcal{L}_{\text{REG}}}
\newcommand{\lwreg}{\mathcal{L}_{\text{W/O-REG}}}

\def\vc{{\bm{c}}}

\def\vy{{\bm{y}}}
\def\vz{{\bm{z}}}

\def\mY{{\bm{Y}}}
\def\mZ{{\bm{Z}}}

\journal{Pattern Recognition}

\begin{document}

\begin{frontmatter}

\title{Multi-Label Contrastive Learning: \\A Comprehensive Study}

\author{Alexandre Audibert\fnref{eq}}
\author{Aur\'elien Gauffre\fnref{eq}}
\author{Massih-Reza Amini}
\fntext[eq]{These authors contributed equally to this work.}

\affiliation{email={Firstname.Lastname@univ-grenoble-alpes.fr\\},
             organization={Univ. Grenoble Alpes, CNRS, Grenoble Computer Science Laboratory (LIG)},
            city={Grenoble},
            postcode={38000},
            country={France}}

\begin{abstract}
Multi-label classification is central to many applications in computer vision and natural language processing. The loss function plays a decisive role in training deep networks for this task, yet conventional losses maximize a likelihood under a label-independence assumption and therefore struggle to capture complex label relationships. Supervised contrastive learning offers a promising alternative, but its extension to the multi-label setting raises specific difficulties related to the definition of positive pairs, label interactions, and long-tailed label distributions. In this paper, we present a comprehensive study of contrastive losses for multi-label classification across diverse settings. Our results show that the gains of contrastive learning stem not only from modeling label interactions but also from the favorable optimization behavior of the contrastive objective. We further identify its limitations: supervised contrastive losses are less effective on datasets with few labels and on ranking-based metrics, whereas they excel, particularly in Macro-F1, on datasets with many labels. Finally, building on a gradient analysis of the multi-label contrastive loss, we propose a regularized loss that cancels the gradient components that cause positive pairs to repel each other. The resulting loss is competitive with or outperforms existing contrastive and non-contrastive losses, with the largest gains in Macro-F1 and in low-data regimes.

\end{abstract}



\end{frontmatter}

\section{Introduction}
Multi-label classification (\MLC{}), which associates each instance with a set of labels, is a fundamental paradigm in machine learning \cite{zhang2014review, liu2022emerging}. It underlies applications ranging from scene understanding and medical image analysis in computer vision (\CV{}) \cite{query2label} to news categorization and legal or medical document classification in natural language processing (\NLP{}) \cite{encoder-decoder, legal-mlc}. Compared with multi-class classification, \MLC{} raises specific challenges: labels follow long-tailed distributions, each instance has far more negative than positive labels, the number of labels varies across instances, and labels are sparse and correlated.

Many approaches have been proposed to model label interactions, based on graphs \cite{gcn-1, gcn-2, gcn-3}, sequential label generation \cite{sgm, gen-order-less}, dedicated regularization terms \cite{laco, effectivemltcco}, label hierarchies \cite{hier-1, hier-2}, label-specific attention \cite{query2label, lsan, ml-decoder}, or vision-language models \cite{dualcoop}. In terms of training objective, the binary cross-entropy (\BCE{}) loss remains the standard choice. It predicts each label independently, and its many variants either reshape its gradient dynamics \cite{focal, asymmetric, dbloss} or adopt a ranking perspective \cite{zlpr}.

Supervised contrastive learning offers an alternative. Instead of acting on the logits, it shapes the representation space directly and accounts for label co-occurrence through interactions between instances. The encoder is first trained with a contrastive objective, and a linear classifier is then trained on the frozen representations. Its application to \MLC{} has long been hindered by long-tailed label distributions and by the difficulty of defining positive pairs when label sets only partially overlap \cite{exploringcontrastivelearninglongtailed}. Recent works address these issues by re-weighting positive pairs according to label frequency \cite{exploringcontrastivelearninglongtailed, mulsupcon} and by introducing label prototypes, which allow contrastive losses to outperform \BCE{} \cite{proto-only}.

The mechanisms behind these gains, however, remain poorly understood. In particular, it is unclear whether they come from instance interactions or from the prototypes, and how contrastive losses behave in low-data regimes, where they are highly effective in the multi-class setting \cite{simple-cl}. The main contributions of this work are as follows:

\begin{enumerate}
    \item \textbf{A comprehensive study of contrastive losses} for multi-label classification in computer vision and natural language processing, covering training from scratch, datasets with few labels, reduced training sets, and full datasets.

    \item \textbf{Insights into their effectiveness}: we show that the gains of contrastive learning in \MLC{} stem not only from modeling label interactions but also from the favorable optimization properties of contrastive objectives.

    \item \textbf{An analysis of their limitations}: the advantage of contrastive losses is most pronounced on datasets with many labels, and it shrinks on datasets with few labels and on ranking-based metrics.

    \item \textbf{A regularized multi-label contrastive loss}: building on a gradient analysis, we propose a regularization term that removes the gradient components pushing positive pairs apart. The resulting loss is competitive with or outperforms existing losses on six benchmarks, with the largest gains in Macro-F1 and in low-data regimes.
\end{enumerate}

The paper is organized as follows. Section~\ref{sec:RW} reviews related work and Section~\ref{sec:MLCL} presents the main multi-label contrastive losses. Section~\ref{sec:method} introduces our gradient analysis and the proposed loss. Sections~\ref{sec:Exps} and~\ref{sec:results} describe the experimental setup and the results, Section~\ref{sec:analysis} analyzes the learned representations and the regularization term, and Section~\ref{sec:conclusion} concludes the paper.

\section{Related Work}
\label{sec:RW}

\subsection{Supervised Loss Functions for Multi-Label Classification}

Binary cross-entropy is the most widely used loss for multi-label classification and provides a probability score for each label. Many variants modify its gradient properties. The \textit{focal loss} \cite{focal} down-weights well-classified instances to focus on hard ones, which benefits the rare labels of long-tailed distributions. Since most labels are absent for any given instance, gradients are dominated by negative labels; the \textit{asymmetric loss} \cite{asymmetric} therefore applies distinct focusing factors to positive and negative labels and discards the gradient of easy negatives. Other losses re-balance labels according to their frequency \cite{dbloss} or jointly exploit sample-wise and label-wise relations \cite{twoway}. A second line of work targets label ranking: the \textit{Log-Sum-Exp Pairwise} (\LSEP{}) loss \cite{lsep} penalizes negative logits that approach or exceed positive ones, and \ZLPR{} \cite{zlpr} adds a fixed threshold separating positive from negative logits, so that the presence or absence of each label can be decided directly.

\subsection{Supervised Contrastive Learning}
Contrastive learning was first developed for self-supervised representation learning, by pulling an anchor toward augmented views of itself and pushing it away from other instances \cite{moco,simple-cl}. \textit{SupCon} \cite{supcon} extends it to the supervised setting by treating instances of the same class as positives. Under balanced conditions, \textit{SupCon} and cross-entropy converge to similar representations, but \textit{SupCon} tends to generalize better thanks to more compact classes \cite{dissecting}. Originally proposed in computer vision, supervised contrastive learning has also proven effective in \NLP{} \cite{negatives-nlp, label-anchor}. On long-tailed data, which distort the representation space in favor of majority classes, it has been improved through re-weighting \cite{balancedcl} and class prototypes acting as stable anchors \cite{paco}.

\subsection{Supervised Contrastive Learning for Multi-Label Classification}

Several works combine supervised contrastive losses with \BCE{} to improve text representations \cite{effectivemltcco, knnmltc}, or use them alongside other objectives, as in a variational approach that aligns label and feature embeddings \cite{MPVAE}, which makes the contribution of the contrastive term hard to isolate. Others rely on label hierarchies to define similarities between instances \cite{hierarchicalContrastiveLearning, hiermltcl, instances}, or on attention mechanisms that learn a representation per label and contrast representations of the same label \cite{dao, instances}.

A central question is how to define positive pairs when label sets only partially overlap. MulSupCon \cite{mulsupcon} treats each instance as a separate sample for each of its labels and averages the \textit{SupCon} loss over labels. Building on a gradient analysis of MulSupCon, Lin and Chen \cite{MICCAI-2025} stratify samples into single- and multi-label cases and weight the loss by class frequency and label co-occurrence to better recognize rare classes in medical images. Other losses grade each pair by the overlap of label sets, for instance through the Jaccard similarity \cite{exploringcontrastivelearninglongtailed, effectivemltcco}. Graded pairs have also been used beyond classification, with soft targets derived from label similarities for Earth observation pretraining \cite{IEEETGRS-2024}, and with positive, soft, and negative pairs for cross-modal hashing under noisy labels \cite{AAAI26}.

In contrast, we study contrastive losses that use a single representation shared by all labels, without hierarchical assumptions or additional standard losses. Our label-overlap re-weighting (Section~\ref{sec:labeldiff}) also grades positive pairs, but our main contribution concerns the optimization dynamics of these losses, namely the repulsive gradients that can arise between positive pairs, which is complementary to the choice of pair weights.

\section{Multi-Label Contrastive Learning}
\label{sec:MLCL}
This section first outlines the motivation for and the main challenges of multi-label contrastive learning. We then introduce our notation, present the state-of-the-art multi-label contrastive losses, and summarize the key insights from previous work.

\subsection{Motivation and Challenges}
Applying contrastive learning to multi-label classification is an appealing research direction. Contrastive learning operates directly on hidden representations and can capture label interactions through the relationships between instances. Extending standard supervised contrastive learning \cite{supcon} to the multi-label setting, however, raises significant challenges. The main difficulty lies in defining positive pairs: two instances may share some labels but not others, so that the notion of similarity becomes graded rather than binary. In addition, labels often follow a long-tailed distribution in which a few labels are associated with many instances while most labels are associated with only a few. This imbalance, already problematic in the multi-class setting, is even more pronounced in the multi-label case, where tail labels frequently co-occur with head labels whose signal can overshadow theirs.

\subsection{Notation}
In the following, let $B$ denote the set of indices of the
examples in a batch. The hidden representation of the $i^{th}$ instance in a batch is denoted by $\vz_{i}$ and $\mathcal{B} = \{\vz_i \mid i \in B \}$ denotes the set of hidden representations of the examples in the batch. Let $L$ denote the total number of labels. Each example $i$ is associated with a label vector $\vy_i \in \{0, 1\}^L$, where the $j^\text{th}$ element, $y_i^j$, indicates the presence ($1$) or absence ($0$) of the $j^\text{th}$ label. We denote by $\Delta(\vz_i) = \{ k\in [1,L] \mid y_i^k = 1\}$ the set of labels of example $i$, and by $P(j, i)~=~\{\vz_l \in \mathcal{B} \mid y_l^j = 1\} \backslash\{\vz_i\}$ the set of representations of the examples having label $j$, excluding that of example $i$. Finally, we define a set of $L$ trainable label prototypes, $C = \{\vc_i\mid i \in \{1,\ldots, L\}\}$.
For ease of presentation, we write the contrastive losses and their derivatives with the dot product instead of the tempered cosine similarity. In practice, however, $\vz_i \cdot \vz_j = \frac{\langle \vz_i, \vz_j \rangle}{\tau \lVert \vz_i \rVert \lVert \vz_j \rVert}$, where $\tau$ denotes the temperature.

\subsection{State-of-the-Art Multi-Label Contrastive Losses}
One of the first losses proposed for {\MLC}, denoted by $\mathcal{L}_{Base}$, is a simple extension of the {\it SupCon} loss \cite{supcon} that weights each pair by the Jaccard similarity of their label sets to model label interactions \cite{exploringcontrastivelearninglongtailed}. 

\begin{equation}
   \mathcal{L}_{Base} = - \frac{1}{|B|} \sum_{i \in B} \frac{1}{Z(i)} \sum_{j \in B \backslash i} \frac{|\vy_i \cap \vy_j|}{|\vy_i \cup \vy_j|} \log\left( \frac{\exp(\vz_i \cdot \vz_j /\tau)}{\sum_{k \in B \backslash i} \exp(\vz_i \cdot \vz_k /\tau)} \right),
    \label{eq:lbase}
\end{equation}

where $Z(i)$ is the normalization term defined as:
\begin{equation*}
   Z(i) = \sum_{j \in B \backslash i} \frac{|\vy_i \cap \vy_j|}{|\vy_i \cup \vy_j|}.
\end{equation*}

This loss is similar to the loss $\mathcal{L}_{JSCL}$ introduced in \cite{effectivemltcco}, the main difference being that the Jaccard weight is placed outside the logarithm. Inside the logarithm, this coefficient would have no effect on training, since $\log(ax)$ and $\log(x)$ have the same derivative.
It is also close to the contrastive loss proposed in \cite{knnmltc}, which uses the dot product between label vectors instead of the Jaccard similarity.

The authors of \cite{proto-only} introduce a loss that relies exclusively on the interactions between instance representations and label prototypes to recognize the types of educational content in videos. Because it forms pairs only with prototypes, this contrastive loss closely resembles the standard cross-entropy loss \cite{proto-anchor,equivalence}, with the prototypes playing the role of the weight matrix of the final layer. Its effectiveness raises the question of whether the performance gains of contrastive learning come from its optimization properties or from the interactions between instances.

We denote this loss by $\mathcal{L}_{Proto}$:
\begin{equation}
    \mathcal{L}_{Proto} = - \frac{1}{|B|} \sum_{i \in B} \frac{1}{|\Delta(\vz_i)|}  \sum_{j \in \Delta(\vz_i)} \log\frac{\exp(\vz_i \cdot \vc_j )}{\sum_{\vc_k \in C} \exp(\vz_i \cdot \vc_k )}.
    \label{eq:lproto}
 \end{equation}
Note that the original paper does not evaluate the learned representations with the standard protocol. Following \textit{SupCon} \cite{supcon}, we slightly modify this loss so that the denominator includes all pairwise interactions. This adjustment serves two purposes: it ensures consistency with the \textit{SupCon} loss, and, by including the numerator term in the denominator, it yields a properly normalized probability distribution.

The MulSupCon loss \cite{mulsupcon} has proven effective for transfer learning in computer vision. Unlike previous works, it treats each instance as a separate sample for each of its labels, and it outperforms more complex approaches such as \cite{MPVAE}.
\begin{equation}
    \mathcal{L}_{MulSupCon} = -\frac{1}{\sum_{i \in B} |\Delta(\vz_i)|} \sum_{i \in B} \ell_{MulSupCon}(\vz_i),
\end{equation}
where $\ell_{MulSupCon}(\vz_i)$ is the individual loss for example $i$ defined as:

\begin{equation}
    \ell_{MulSupCon}(\vz_i) = \sum_{j \in \Delta(\vz_i)} \frac{1}{|P(j,i)|}  \sum_{\vz_l \in P(j,i)} \log \frac{\exp(\vz_i \cdot \vz_l )}{\sum_{ k \in B \backslash \vz_i}\exp(\vz_i \cdot \vz_k)} 
\end{equation}

The Multi-label Supervised Contrastive (\MSC{}) loss \cite{exploringcontrastivelearninglongtailed} has proven effective in addressing long-tailed label distributions in \NLP{} tasks. It corrects the \textit{attraction-repulsion imbalance} through a re-weighting scheme based on label frequency, and it mitigates the \textit{lack of positives} with a MoCo \cite{moco} queue and label prototypes:
\begin{equation*}
\mathcal{L}_{MSC}\!=\! -\frac{1}{|B|} \sum_{i \in B}  \frac{1}{|\vy_i|} \!\!\sum_{j \in \Delta(\vz_i)}\frac{1}{N(i, j)} 
\!\!\sum_{\vz_l \in P(j,i)\cup \vc_j }\!\!\!\!\! f(\vz_i, \vz_j) \log \frac{e^{\vz_i \cdot \vz_l /\tau}}{\!\!\displaystyle \sum_{\vz_k\in \mathcal{B} \cup C \backslash \vz_i}\!\!\!g_i(\vz_k, \beta)e^{\vz_i \cdot \vz_k/\tau}}.
\end{equation*}
The functions $f$ (as defined in Eq.~\ref{eq:f}) and $g$ (Eq.~\ref{eq:g}) re-weight the attraction term so that the most frequent labels do not dominate the optimization and less frequent labels receive balanced attention.

\begin{equation}
f(\vz_i, \vz_j) =
\begin{cases}
    1 & \text{if } \vz_j \in C, \\
    \frac{1}{|\vy_i \cup \vy_j|} & \text{otherwise.}
\end{cases}
\label{eq:f}
\end{equation}
\begin{equation}
    g_i(\vz_k, \beta) =
\begin{cases}
    1 & \text{if } \vz_k \in C, \\
    \beta & \text{otherwise. }
\end{cases}
\label{eq:g}
\end{equation}
A major difference between the \MSC{} loss \cite{exploringcontrastivelearninglongtailed} and the MulSupCon loss \cite{mulsupcon} is that the former assigns the same weight to every instance, whereas the latter weights each instance in proportion to its number of labels.

\subsection{Insights from Previous Work}
In summary, recent approaches rely on two key components: label prototypes \cite{proto-only, exploringcontrastivelearninglongtailed} and the weighting of positive pairs. A common strategy for weighting positive pairs is to account for label frequency, either by treating each instance as a separate sample for each of its labels \cite{mulsupcon} or by re-normalizing the weights according to label frequency within the batch \cite{exploringcontrastivelearninglongtailed}.
\section{The Proposed Multi-Label Contrastive Loss}
\label{sec:method}
In this section, we introduce a new multi-label contrastive loss motivated by the insights from previous work and by a gradient analysis of existing losses.

\subsection{Gradient Analysis}

We first examine the gradient of the multi-label contrastive losses described above, using the standard notation of \textit{SupCon} \cite{supcon}: $P(i)$ denotes the set of indices of the positive instances of anchor $\vz_i$, $N(i)$ the set of indices of its negative instances, and $A(i)$ the set of indices of the batch excluding $i$. The definition of positive and negative pairs depends on the specific loss considered.

All the multi-label contrastive losses above, such as $\mathcal{L}_{Base}$, $\mathcal{L}_{Proto}$, and $\mathcal{L}_{MSC}$, can be written in the following general form, with the exception of MulSupCon \cite{mulsupcon} because of its different normalization scheme:

\begin{equation}
    \begin{aligned}
        &\mathcal{L}_{contrastive} = -\frac{1}{|B|} \sum_{i \in B} \ell(\vz_i),\\
        &\ell(\vz_i) = \frac{1}{\sum_{p \in P(i)}  \lambda_p^i} \sum_{j \in P(i)} \lambda_j^i \log \frac{\exp(\vz_j \cdot \vz_i)}{\sum_{k \in A(i) } \exp(\vz_k \cdot \vz_i)}.
    \end{aligned}
    \label{eq:contrastive-general}
\end{equation}

This formulation generalizes the \textit{SupCon} loss \cite{supcon}: $\lambda_j^i$ represents the target similarity between the anchor $\vz_i$ and its positive instance $\vz_j$, and \textit{SupCon} corresponds to $\lambda_j^i = 1$.

The gradient of $\ell(\vz_i)$ with respect to the anchor $\vz_i$ is given in Eq.~\eqref{eq:deriveative-anch}, and its gradient with respect to $\vz_k$ is given in Eq.~\eqref{eq:deriveative-pos} for a positive pair and in Eq.~\eqref{eq:deriveative-neg} for a negative pair. 
\begin{align}
    \frac{\partial \ell(\vz_i)}{\partial \vz_i} &=  \sum_{j \in P(i)} \left( \frac{\lambda_j^i}{\sum_{p \in P(i)}  \lambda_p^i}   -\sigma_{j, i}\right) \vz_j  - \sum_{n \in N(i) } \sigma_{n, i} \vz_n, \label{eq:deriveative-anch}\\
    \frac{\partial \ell(\vz_i)}{\partial \vz_k} &=  \left( \frac{ \lambda_k^i}{ \sum_{p \in P(i)} \lambda_p^i} - \sigma_{k, i}\right) \vz_i \label{eq:deriveative-pos} ~~~\text{if}, ~~ k \in P(i),\\
    \frac{\partial \ell(\vz_i)}{\partial \vz_k} &=   -\sigma_{k, i} \vz_i ~~~\text{if}, ~~ k \in N(i),\label{eq:deriveative-neg} 
\end{align}

where $\sigma_{j,i} = \frac{ \exp(\vz_j \cdot \vz_i) }{\sum_{k \in A(i) } \exp(\vz_k \cdot \vz_i)}$ is the normalized similarity between $\vz_j$ and $\vz_i$, a key quantity in the following analysis.
Note that, if $\sigma_{k, i}$ is sufficiently high, the coefficient $\frac{ \lambda_k^i}{ \sum_{p \in P(i)} \lambda_p^i} - \sigma_{k, i}$ in Eq. \eqref{eq:deriveative-pos} becomes negative, so the gradient of the positive pair has the same sign as that of a negative pair (Eq. \eqref{eq:deriveative-neg}): the two positive instances are pushed apart.

The contrastive loss reaches its minimum when all these gradients vanish, that is, when
\begin{equation}
    \sigma_{k,i} = 
    \begin{cases}
        \frac{\lambda_k^i}{\sum_{p \in P(i)}  \lambda_p^i} & \text{if} ~~ k \in P(i), \\
        0 & \text{if} ~~ k \in N(i). \\
    \end{cases}
    \label{eq:contrastive-learning-minimum}
\end{equation}
This highlights the importance of properly defining both the target scores of positive pairs and the set of negative pairs: the effectiveness of multi-label contrastive learning is highly sensitive to the choice of the weighting coefficients $\lambda_j^i$.

The gradients in Eqs.~\eqref{eq:deriveative-anch}--\eqref{eq:deriveative-neg} further reveal that, under certain conditions, the gradient of a positive pair can mirror that of a negative pair, which may hinder optimization.
Indeed, Eq.~\eqref{eq:deriveative-neg} shows that the gradient of a negative pair points in the direction of its representation $\vz_k$.
For a positive pair, Eq.~\eqref{eq:deriveative-pos} shows that if the score $\sigma_{k, i}$ is sufficiently high, i.e., if the two embeddings are already very close, the coefficient $\left(- \frac{\lambda_k^i}{\sum_{p \in P(i)} \lambda_p^i} + \sigma_{k, i}\right)$ becomes positive.
In this case, the gradient of the positive pair also points in the direction of $\vz_k$ and behaves like that of a negative pair: the two positive instances are pushed apart. The same observation holds for the contributions of positive and negative pairs to the gradient with respect to the anchor in Eq.~\eqref{eq:deriveative-anch}.

\subsection{Gradient Regularization}

To counter this undesirable behavior, we introduce a regularization term $\ell_{reg}(\vz_i)$ that corrects the gradient whenever needed, i.e., when $-\Lambda_j^i + \sigma_{j, i} > 0$, where the normalized coefficient $\Lambda_j^i \in \mathbb{R}$ is defined as:
\begin{equation}
\Lambda_j^i = \frac{\lambda_j^i}{\sum_{p \in P(i)} \lambda_p^i}. 
\end{equation}

The regularization term $\ell_{reg}(\vz_i)$ is defined as follows:
\begin{equation}
    \ell_{reg}(\vz_i) = \sum_{j \in P(i)} \max(0, (-\Lambda_j^i + \sigma_{j, i}.detach())) \vz_j \cdot \vz_i.
\end{equation}
where the $detach()$ operator stops the gradient from flowing through the coefficient.

This term rebalances the gradient so that positive pairs never behave like negative pairs. Its gradients with respect to the anchor $\vz_i$ and to a positive instance $\vz_k$ are given in Eqs.~\eqref{eq:derivative-reg-achor} and~\eqref{eq:derivative-reg-positive}.
\begin{equation}
    \begin{aligned}
        \frac{\partial \ell_{reg}(\vz_i)}{\partial \vz_i} &=  \sum_{j \in P(i)} 
            \mathbbm{1}_{(-\Lambda_j^i + \sigma_{j, i}) > 0 }
            (-\Lambda_j^i + \sigma_{j, i}) \vz_j. \label{eq:derivative-reg-achor}
    \end{aligned}
\end{equation}
\begin{equation}
    \begin{aligned}
        \frac{\partial \ell_{reg}(\vz_i)}{\partial \vz_k} &= 
            \mathbbm{1}_{(-\Lambda_k^i + \sigma_{k, i}) > 0}
            (-\Lambda_k^i + \sigma_{k, i}) \vz_i. \label{eq:derivative-reg-positive}
    \end{aligned}
\end{equation}
The regularized per-anchor loss $\hat{\ell}(\vz_i)$ then replaces $\ell(\vz_i)$ in Eq.~\eqref{eq:contrastive-general}:
\begin{equation}
    \hat{\ell}(\vz_i) = \ell(\vz_i) + \ell_{reg}(\vz_i).
\end{equation}

\begin{figure}[h!]
    \centering
    \includegraphics[width=.65\textwidth]{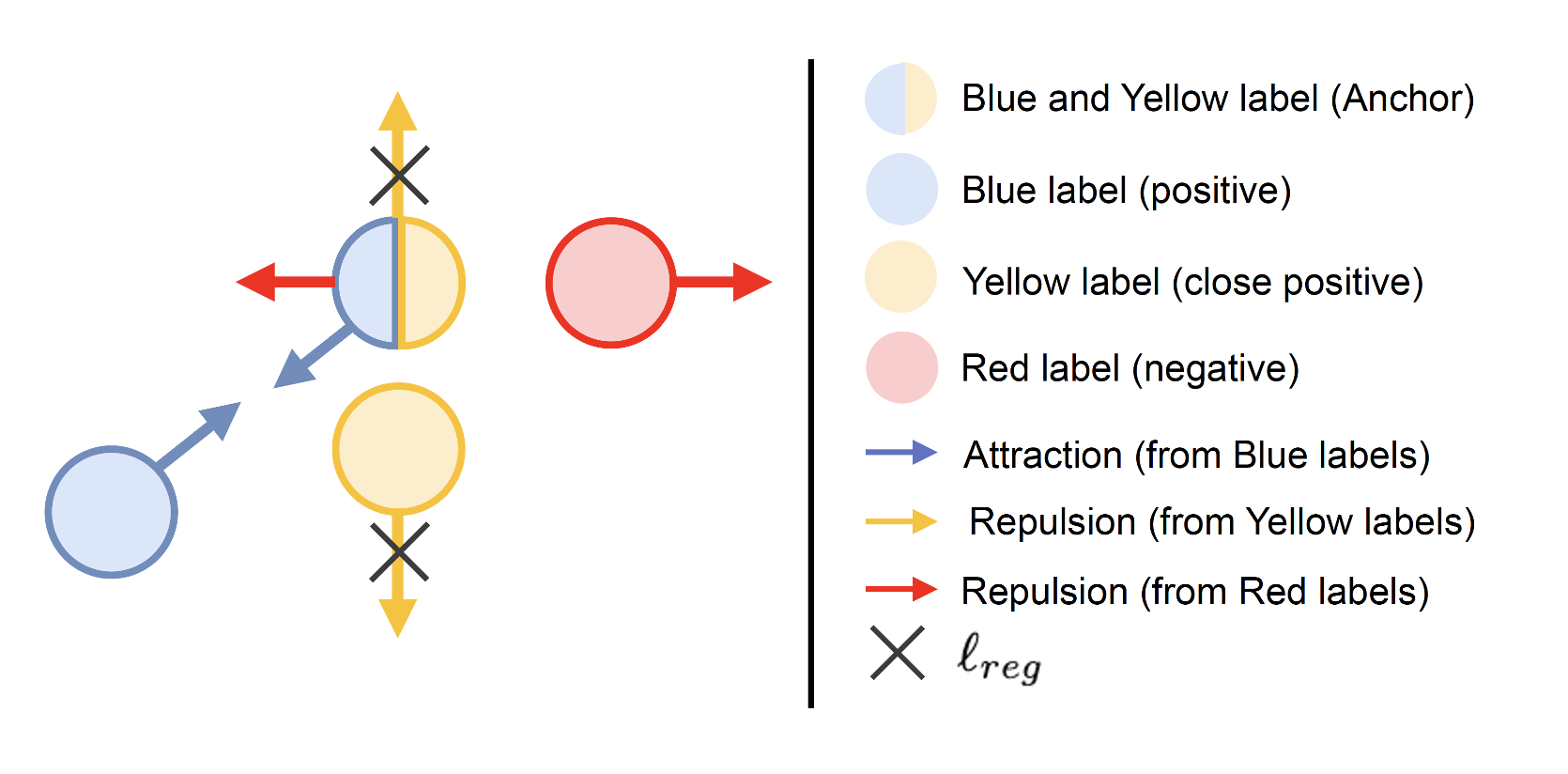}
    \caption{Gradient behavior of the multi-label contrastive loss. Embeddings sharing the yellow label that are already very close generate gradients that cause unintended repulsion. The proposed regularization term cancels this repulsion and preserves the attraction between similar examples.}
    \label{tab:reg}
\end{figure}

As illustrated in Fig.~\ref{tab:reg}, the regularization term prevents positive pairs that are already very close from generating repulsive gradients. Notably, $\hat{\ell}(\vz_i)$ and $\ell(\vz_i)$ share the same minimum, given by Eq.~\eqref{eq:contrastive-learning-minimum}, since the regularization term is inactive at this point. Any performance improvement therefore stems solely from a different optimization behavior. 

\subsection{Regularized Multi-Label Contrastive Loss}

We now design a simple contrastive loss that combines the regularization term $\ell_{reg}$ with the insights from previous work.
Following \cite{mulsupcon, exploringcontrastivelearninglongtailed}, we normalize the contribution of positive pairs according to label frequency within the batch and incorporate label prototypes, treating all prototypes in $C$ as members of the batch $\mathcal{B}$. As in contrastive semi-supervised learning \cite{proto-anchor}, this allows the prototypes to act as stable reference points for each label.

Our regularized multi-label contrastive loss, denoted by $\lreg$, is defined as:

\begin{equation}
    \lreg = - \frac{1}{|B|} \sum_{i \in B } \hat{\ell}(\vz_i),
    \label{eq:main_loss}
\end{equation}
where,
\begin{equation}
    \hat{\ell}(\vz_i) = \frac{1}{|\vy_i|} \sum_{j \in \Delta(\vz_i)} \frac{1}{|P(j,i)|} 
    \sum_{\vz_l \in P(j,i)} \log  \frac{\exp(\vz_i \cdot \vz_l)}{\sum_{k \in B \setminus i} \exp(\vz_i \cdot \vz_k)} + \ell_{reg}(\vz_i).
    \label{eq:sub_loss}
\end{equation}

\subsection{Accounting for Label Differences Between the Anchor and Its Positives}
\label{sec:labeldiff}
As noted in \cite{mulsupcon}, the attraction between two instances is proportional to the number of labels they share.
A positive instance may, however, also carry labels that are unrelated to the anchor. Some works account for this through the Jaccard similarity \cite{exploringcontrastivelearninglongtailed, effectivemltcco}.
To reflect varying degrees of label overlap, we propose to re-weight each positive pair by the fraction of its labels shared with the anchor:

\begin{equation}
    f(\vy_i, \vy_j) = \left(\frac{|\vy_i \cap \vy_j|}{|\vy_j|}\right)^{\alpha},
    \label{eq:reg-label-diff}
\end{equation}
where $\alpha \in \mathbb{R}^{+}$ controls the influence of label overlap; larger values of \(\alpha\) penalize label differences more strongly.
With this re-weighting, the per-anchor loss becomes
\begin{equation}
        \hat{\ell}(\vz_i) = \frac{1}{|\vy_i|}\sum_{j \in \Delta(\vz_i)} \frac{1}{Z(j,i)}  \sum_{\vz_l \in P(j,i)} f(\vy_i, \vy_l) \log \frac{\exp(\vz_i \cdot \vz_l )}{\sum_{k\in B \backslash i}\exp(\vz_i \cdot \vz_k)} + \ell_{reg}(\vz_i),
    \label{eq:simple}
\end{equation}

where $Z(j,i)$ is the normalization term:
\begin{equation}
    Z(j,i) = \sum_{\vz_l \in P(j,i)} f(\vy_i, \vy_l).
\end{equation}

\section{Experimental Setup}
\label{sec:Exps}
This section describes the datasets used in our experiments, the baseline losses against which we compare the proposed regularized multi-label contrastive loss, and the implementation details.

\subsection{Datasets}
To assess the generality of the studied losses, we conduct experiments in both the \CV{} and \NLP{} domains. For each dataset, we also build a subset with limited training data to evaluate the different approaches in a low-data setting.

\subsubsection{Computer Vision}
\begin{itemize}
    \item \textbf{PASCAL VOC} \cite{pascal} contains images annotated with multiple object classes among 20 categories.
    \item \textbf{MS-COCO} \cite{coco} (Microsoft Common Objects in Context, \COCO{}) contains images of a wide range of everyday scenes, annotated with 80 object categories.
    \item \textbf{NUS-WIDE} \cite{nuswide} consists of Flickr images annotated with tags from a set of 81 concepts.
\end{itemize}
\subsubsection{Natural Language Processing}
\begin{itemize} \item \textbf{RCV1-v2} \cite{rcv1-v2} consists of newswire stories from Reuters Ltd., each of which can be assigned to several of the 103 original topics. We keep the original training/test split but modify the number of labels (see Table~\ref{table:datasets:1}). \item \textbf{AAPD} \cite{sgm} (Arxiv Academic Paper Dataset) contains the abstracts and subjects of 55,840 academic papers, each of which may belong to several subjects. \item \textbf{BGC} \cite{bgc} (Blurb Genre Collection) is a multi-label dataset of promotional descriptions of books annotated with their genres. \end{itemize}

Since the image datasets do not provide publicly labeled test sets, we train on the original training split and divide the original validation set equally into a validation set and a test set. This prevents data leakage and ensures that the test set remains entirely unseen during training and model selection (in particular during the small grid search performed for linear evaluation), leading to a more reliable evaluation.
Table~\ref{table:datasets:1} summarizes the characteristics of the datasets.

\begin{table}[h!]
\begin{center}
\caption{Dataset statistics: number of examples (in thousands) in the training, validation, and test sets, number of labels $L$, and average number of labels per example $\overline{L}$. ``S'' denotes the subsets used in the low-data experiments.}
\resizebox{\textwidth}{!}{\begin{tabular}{clcccccccc}
\toprule[1pt] 
& Dataset & $|\text{Train}|$ & $|\text{Val}|$ & $|\text{Test}|$ & $|\text{Train S}|$ & $|\text{Val S}|$ & $|\text{Test S}|$ & L & $\overline{L}$ \\
\midrule
\parbox[t]{2mm}{\multirow{3}{*}{\rotatebox[origin=c]{90}{\CV}}} & PASCAL \cite{pascal} &     5.0k &     2.5k &     2.5k &                1.0k &     2.5k &     2.5k &       20 &      1.5 \\
   &  COCO \cite{coco} &    82.0k &    20.2k &    20.2k &               16.4k &    20.2k &    20.2k &       80 &      2.9 \\
   & NUS-WIDE \cite{nuswide} &   125.4k &    41.9k &    41.9k & 25.1k &    41.9k &    41.9k &       81 &      2.4 \\

\midrule

\parbox[t]{2mm}{\multirow{3}{*}{\rotatebox[origin=c]{90}{\NLP}}}  & RCV1 \cite{rcv1-v2}  & 19.6k & 3.4k & 781.2k & 4.9k & 10.8k & 788.6k & 91 & 3.2 \\
 & AAPD \cite{sgm} & 37.8k & 6.7k & 11.3k & 4.3k & 16.3k & 35.1k & 54 & 2.4 \\
 & BGC \cite{bgc}  & 58.7k & 14.7k & 18.3k & 5.8k & 30.6k & 55.3k & 146 & 3.0 \\
\hline
\end{tabular}}
\label{table:datasets:1}

\end{center}

\end{table}

\subsection{Baselines}
In addition to the contrastive losses presented in Section~\ref{sec:MLCL}, we compare against the following standard supervised losses for multi-label classification.
\subsubsection{Standard Supervised Losses}

\noindent $\bullet~~ \mathcal{L}_{BCE}$ denotes the binary cross-entropy loss: 
\begin{equation*}
   \mathcal{L}_{BCE} = -\frac{1}{N} \sum_{i=1}^N \frac{1}{L} \sum_{j=1}^L \\
    y_i^j\log( \hat{y}_i^j) + (1 - y_i^j)\log( 1 - \hat{y}_i^j),
    \label{eq:bce}
\end{equation*}
where $\hat{\boldsymbol{y}}_i=(\hat{y}_i^1, \ldots, \hat{y}_i^L)$ is the vector of output probabilities of the model for instance $i$.

\noindent $\bullet~~ \mathcal{L}_{ASY}$ denotes the asymmetric loss \cite{asymmetric}:

\begin{equation}
    \mathcal{L}_{ASY} = - \frac{1}{N} \sum_{i=1}^N \frac{1}{L} \sum_{j=1}^L y_i^j(1 - s_i^j)^{\gamma^+}\log(s_i^j)  + (1 - y_i^j)(s_i^j)^{\gamma^-}\log(1 - s_i^j),
\end{equation}
with $s_i^j = \max(\hat{y}_i^j - m, 0)$, where $m$ is the hard threshold and $\gamma^+$ and $\gamma^-$ are the focusing parameters for positive and negative labels.

\noindent $\bullet~~ \mathcal{L}_{ZLPR}$ denotes the \textit{Zero-bounded Log-sum-exp and Pairwise Rank-based} loss \cite{zlpr}:
\begin{equation}
    \mathcal{L}_{ZLPR} = \frac{1}{N} \sum_{i=1}^N \left[
    \log\Big(1 + \sum_{j=1}^L y_i^j\, e^{-o_i^j}\Big)
    + \log\Big(1 + \sum_{j=1}^L (1 - y_i^j)\, e^{o_i^j}\Big) \right],
    \label{eq:zlpr_opt}
\end{equation}
where $o_i^j \in \mathbb{R}$ denotes the logit (pre-sigmoid output) of the model for
label $j$ of instance $i$, and $\hat{y}_i^j = \sigma(o_i^j)$ the corresponding probability.
\subsection{Implementation Details}
The source code of our loss and of all the losses used for comparison is publicly available on GitHub\footnote{\url{https://github.com/audibert-alexandre-fra/Multi-label-contrastive-comprehensive}}, and the pseudocode of our loss is given in \ref{appendix:pseudocode}.
In all experiments, we apply gradient clipping with a maximum norm of 1 and use 16-bit automatic mixed precision to reduce training time and memory consumption.

For the standard supervised losses, we use a linear learning rate scheduler with a 5\% warm-up. During training, we retain the checkpoint with the best validation Micro-F1 score, and we select the hyperparameter configuration with the best validation results averaged over several seeds for testing.
For the asymmetric loss \cite{asymmetric}, we set $\gamma^+ = 0$, $\gamma^- = 1$, and $m = 0$, following common practice \cite{query2label}. 

For contrastive learning, following \textit{SupCon} \cite{supcon}, we use a cosine learning rate scheduler with a 5\% warm-up and set the temperature $\tau$ to 0.1. The projection head consists of two fully connected layers with a ReLU activation, \( W_2 \cdot \text{ReLU}(W_1 \cdot x) \), where \( W_1 \in \mathbb{R}^{h \times h} \) and \( W_2 \in \mathbb{R}^{d \times h} \), where \( h \) is the dimension of the hidden space and \( d = 256 \). For evaluation, we keep the last checkpoint, discard the projection head, and train one logistic regression classifier per label with the binary cross-entropy loss on the frozen representations.

Results are reported on the test set with standard metrics, namely the Hamming loss, the Micro-F1 score, and the Macro-F1 score \cite{exploringcontrastivelearninglongtailed}.

\subsubsection{Natural Language Processing}
For the \NLP{} experiments, we use no data augmentation and truncate documents to 220 tokens. Following previous works \cite{zlpr, instances}, we use an encoder-only model, RoBERTa-base \cite{roberta}, as the backbone, with the Hugging Face implementation\footnote{\url{https://huggingface.co/roberta-base}}. The pre-trained model keeps a dropout rate of 0.1, and the \texttt{[CLS]} token is used as the text representation, followed by a fully connected layer as the decoder. We use the AdamW optimizer \cite{AdamW} with a weight decay of 0.01, from which bias and LayerNorm parameters are exempted, as is common practice.
                                   
\subsubsection{Computer Vision}

Unless otherwise specified, all computer vision experiments use a ResNet-50 \cite{resnet} pre-trained on ImageNet \cite{imagenet}. The default optimizer is stochastic gradient descent (\SGD{}) with momentum, with a batch size of 64 and an input resolution of $224 \times 224$ (dataset-specific optimizers are listed in \ref{Appendix:implementationdetails}).

For data augmentation, we apply RandAugment with standard parameters (\(n=2, m=9\)) for both contrastive training and linear evaluation. According to \cite{supcon}, this configuration performs comparably to their best setting, which used AutoAugment. For a fair comparison, we use the same augmentation strategy for the models trained with \BCE{}-based losses.

For the non-contrastive losses, we train for 5, 10, and 20 epochs and select the best model on the validation set. For the contrastive losses, which converge faster and are less prone to overfitting, we use 10 epochs throughout. Given the impact of data augmentation on contrastive learning, we leave the exploration of alternative augmentation strategies to future work.

In all experiments, we remove the MoCo queue \cite{moco} so that all contrastive losses are compared without additional implementation tricks. Further implementation details are provided in \ref{Appendix:implementationdetails}.
\section{Results}
\label{sec:results}
We first report the results obtained on the full datasets and then those obtained in low-data regimes.
\begin{table*}[]
\centering
\caption{Comparison of multi-label loss functions on the vision datasets (PASCAL, MS-COCO, and NUS-WIDE) and the text datasets (AAPD, RCV1, and BGC). Metrics are Micro-F1 ($\mu$-F$_1$), Macro-F1 (M-F$_1$), and Hamming loss ($\times 10^3$). The best results, and those not significantly different from them, are in bold. All evaluated losses are implemented in our public repository.}
\resizebox{\textwidth}{!}{%
\small
\begin{tabular}{lccc ccc ccc c ccc ccc ccc}
\toprule[1pt] 
\multicolumn{1}{l}{} & \multicolumn{3}{c}{PASCAL} & \multicolumn{3}{c}{MS-COCO} & \multicolumn{3}{c}{NUS-WIDE} & & \multicolumn{3}{c}{AAPD} & \multicolumn{3}{c}{RCV1} & \multicolumn{3}{c}{BGC}\\
\cmidrule(r){2-4} \cmidrule(lr){5-7} \cmidrule(l){8-10}  \cmidrule(r){12-14} \cmidrule(lr){15-17} \cmidrule(l){18-20}
\multicolumn{1}{l}{\textbf{Loss}} & $\mu$-F$_1$ & M-F$_1$ & Ham & $\mu$-F$_1$ & M-F$_1$ & Ham & $\mu$-F$_1$ & M-F$_1$ & Ham & & $\mu$-F$_1$ & M-F$_1$ & Ham & $\mu$-F$_1$ & M-F$_1$ & Ham & $\mu$-F$_1$ & M-F$_1$ & Ham \\
\cmidrule{1-20}
 & \multicolumn{19}{c}{\textbf{Non-Contrastive Losses}} \\
\cmidrule{1-20}
$\mathcal{L}_{\text{BCE}}$ & 83.40 & 80.93 & \textbf{2.41} & 62.23 & 48.17 & 2.13 & 65.18 & 16.08 & 1.72 & & 73.39 & 59.59 & 22.73 & 88.30 & 75.80 & 8.17 & 80.46 & 65.05& 7.85\\
$\mathcal{L}_{\text{ASY}}$ \cite{asymmetric} & \textbf{83.84} & \textbf{81.60} & 2.44 & 65.06 & 52.56 & 2.08 & 65.50 & 14.62 & 1.74 & & 73.51 & 59.78 & 22.99 & 88.12 & 74.72 & 8.33 & 80.49 & 65.67 & 7.91\\
$\mathcal{L}_{\text{ZLPR}}$ \cite{zlpr} & 83.37 & 80.88 & \textbf{2.42} & 70.43 & 63.88 & \textbf{1.83} & 70.47 & 38.02 & 1.61 & & 73.64 & 59.7 & \textbf{22.62} & \textbf{88.52} & 76.48 & \textbf{8.0} & 80.91 & 66.53 & 7.70\\

\cmidrule{1-20}
& \multicolumn{19}{c}{\textbf{Contrastive Losses}} \\
\cmidrule{1-20}
$\mathcal{L}_{\text{Base}}$ \cite{exploringcontrastivelearninglongtailed} & 82.53 & 79.86 & 2.57 & 69.81 & 64.22 & 1.97 & 71.07 & 52.84 & 1.64 &  & 73.53 & 60.13 & 23.39 & 87.95 & 74.89 & 8.58 & 80.63 & 62.78 & 7.85\\
$\mathcal{L}_{\text{Proto}}$ \cite{proto-only} & 81.85 & 79.75 & 2.69 & 71.88 & 67.35 & 1.86 & 71.79 & 56.03 & 1.62 &  & 73.59 & 61.02 & 23.79 & 88.3 & 77.23 & 8.36 & 80.52 & 65.49 & 8.04\\
$\mathcal{L}_{\text{MulSupCon}}$ \cite{mulsupcon} & 82.75 & 80.26 & 2.53 & 71.33 & 66.25 & 1.88 & 71.88 & 54.36 & 1.61 &  & 73.64 & 60.52 & 23.28 & 88.47 & 76.96 & 8.2 & 81.00 & 64.81 & 7.74\\
$\mathcal{L}_{\text{MSC}}$ \cite{exploringcontrastivelearninglongtailed} & 82.56 & 80.39 & 2.58 & 71.96 & 67.54 & 1.85 & \textbf{72.02} & 56.13 & \textbf{1.60} & & \textbf{73.68} & 60.67 & 23.49 & 88.41 & 77.36 & 8.25 & 81.07 & 65.89 & \textbf{7.69}\\
$\lwreg$ & 82.23 & 79.99 & 2.63 & 71.78 & 67.50 & 1.86 & \textbf{71.99} & \textbf{56.59} & \textbf{1.61} & & 73.62 & 60.88 & 23.50 & 88.43 & 77.24 & 8.2 & 81.03 & 66.00 & 7.73\\
$\lreg$ (Eq. \ref{eq:main_loss})& 83.08 & 80.57 & 2.52 & \textbf{72.19} & \textbf{68.03} & \textbf{1.83} & \textbf{72.05} & \textbf{56.52} & \textbf{1.60} & & \textbf{73.72} & \textbf{61.28} & 23.57 & 88.41 & \textbf{77.45} & 8.27 & 80.83 & \textbf{66.66} & 7.84\\
\bottomrule[1pt] 
\end{tabular}%
}

\vspace{2px}
\label{table:results-all}
\end{table*}

\subsection{Standard Supervised Losses}
Among the standard supervised losses, $\mathcal{L}_{ZLPR}$ outperforms both $\mathcal{L}_{ASY}$ and $\mathcal{L}_{BCE}$ on all metrics for the datasets with 80 labels or more (Table~\ref{table:results-all}). This highlights its ability to handle large label spaces with complex label interactions, in contrast to \BCE{}-based losses. A likely explanation is that the ratio of positive to negative labels decreases as the number of labels grows, a setting in which \BCE{}-based losses are known to struggle \cite{asymmetric}. When the number of labels is smaller, as for \PASCAL{} and \AAPD{}, the asymmetric loss performs on par with or better than $\mathcal{L}_{ZLPR}$. This observation is confirmed by additional experiments on a dataset with few labels in \ref{Appendix:semeval}. A comparison of these losses in terms of ranking metrics such as mAP is provided in \ref{Appendix:ranking}.

\subsubsection{Contrastive Losses}Among the contrastive losses, the proposed loss $\lreg$ (Eq.~\eqref{eq:main_loss}) outperforms previous methods on almost all metrics for the image datasets. For the text datasets, the results are more mixed: $\lreg$ still achieves the best Macro-F1 scores but slightly lower Micro-F1 scores.

\subsubsection{Standard Versus Contrastive Losses} Except on \PASCAL{}, the dataset with the fewest labels (20), contrastive losses outperform the standard loss functions. Even simple contrastive losses are competitive with the best non-contrastive losses on MS-COCO, NUS-WIDE, and \AAPD{}.

\ref{appendixB} further examines the behavior of the studied losses when models are trained from scratch, and shows that the proposed loss converges faster and is more resistant to overfitting.

\subsection{Low-Data Regimes}
\begin{table*}[]
\caption{Comparison of multi-label loss functions in the low-data regime, using 20\% of the training examples for the vision datasets (PASCAL, MS-COCO, and NUS-WIDE) and the subsets described in Table~\ref{table:datasets:1} for the text datasets (AAPD, RCV1, and BGC). Metrics are Micro-F1 ($\mu$-F$_1$), Macro-F1 (M-F$_1$), and Hamming loss ($\times 10^3$). The best results, and those not significantly different from them, are in bold.}
\centering
\resizebox{\textwidth}{!}{%
\small
\begin{tabular}{lccc ccc ccc c ccc ccc ccc}
\toprule[1pt] 
\multicolumn{1}{l}{} & \multicolumn{3}{c}{PASCAL} & \multicolumn{3}{c}{MS-COCO} & \multicolumn{3}{c}{NUS-WIDE} & & \multicolumn{3}{c}{AAPD} & \multicolumn{3}{c}{RCV1} & \multicolumn{3}{c}{BGC}\\
\cmidrule(r){2-4} \cmidrule(lr){5-7} \cmidrule(l){8-10} \cmidrule(r){12-14} \cmidrule(lr){15-17} \cmidrule(l){18-20}
\multicolumn{1}{l}{\textbf{Loss}} & $\mu$-F$_1$ & M-F$_1$ & Ham & $\mu$-F$_1$ & M-F$_1$ & Ham & $\mu$-F$_1$ & M-F$_1$ & Ham & & $\mu$-F$_1$ & M-F$_1$ & Ham & $\mu$-F$_1$ & M-F$_1$ & Ham & $\mu$-F$_1$ & M-F$_1$ & Ham \\
\cmidrule{1-20}
& \multicolumn{19}{c}{\textbf{Non-Contrastive Losses}} \\
\cmidrule{1-20}
$\mathcal{L}_{\text{BCE}}$ & 75.27 & 66.17 & 3.32 & 54.07 & 34.10 & 2.46 & 67.38 & 24.57 & 1.69 & & 68.76 & 49.49 & 25.93 & 86.90 & 70.39 & 9.31 & 74.28 & 49.61 & 10.06\\
$\mathcal{L}_{\text{ASY}}$ &  77.03 & 71.04 & 3.28 & 57.22 & 39.59 & 2.43 & 68.69 & 27.43 & \textbf{1.68} & & 68.85 & 51.40 & 27.22 & 86.91 & 72.34 & 9.05 & 74.64 & 52.29 & 10.34\\
$\mathcal{L}_{\text{ZLPR}}$ & 78.73 & 74.96 & \textbf{2.96} & 65.71 & 57.79 & \textbf{2.07} & 69.10 & 43.63 & \textbf{1.68} & & \textbf{69.45} & 50.98 & \textbf{25.69} & \textbf{87.39} & 72.05 & \textbf{8.82} & 75.49 & 53.22 & 9.84\\
\cmidrule{1-20}
& \multicolumn{19}{c}{\textbf{Contrastive Losses}} \\
\cmidrule{1-20}
$\mathcal{L}_{\text{Base}}$  & 79.21 & 76.19 & 3.07 & 64.83 & 58.35 & 2.36 & 68.54 & 47.68 & 1.82 & & 69.42 & 49.85 & 26.13 & 86.62 & 71.03 & 9.53 & 75.26 & 51.75 & 9.87\\
$\mathcal{L}_{\text{Proto}}$  & 78.63 & 75.91 & 3.15 & 66.50 & 60.99 & 2.25 & 69.02 & 49.69 & 1.79 & & 69.09 & 51.05 & 27.79 & 86.52 & 72.91 & 9.76  & 74.82 & 53.38 & 10.52 
\\
$\mathcal{L}_{\text{MulSupCon}}$  & 78.94 & 76.36 & 3.10 & 66.02 & 59.86 & 2.27 & 68.98 & 48.16 & 1.79 & & \textbf{69.39} & 49.01 & \textbf{25.70} & 86.96 & 71.84 & 9.3 & \textbf{75.57} & 51.93 & \textbf{9.77}\\
$\mathcal{L}_{\text{MSC}}$  & 78.83 & 76.59 & 3.17 & 66.34 & 60.65 & 2.26 & 69.19 & 49.64 & 1.78 & & \textbf{69.45} & 50.69 & 26.77 & 86.91 & 72.49 &  9.42 & \textbf{75.63} & 52.57 & \textbf{9.77}\\
$\lwreg$ & 78.59 & 76.16 & 3.20 & 66.46 & 61.06 & 2.25 & 69.37 & 50.19 & 1.75 & & 69.23 & 49.95 & 26.41 & 86.77 & 72.17 &  9.45 & 75.51 & 52.82 & 9.89\\
$\lreg$ & \textbf{79.63} & \textbf{76.77} & 3.02 & \textbf{67.04} & \textbf{61.90} & 2.20 & \textbf{69.63} & \textbf{50.39} & 1.75 & & 68.89 & \textbf{52.30} & 29.11 & 86.5 & \textbf{73.09} &  9.78 & 75.21 & \textbf{53.44} & 10.4\\
\bottomrule[1pt] 
\end{tabular}%
}
\vspace{2px}
\label{table:results-small}
\end{table*}

We next evaluate the losses when training data are scarce, a common situation in practice. The results in Table~\ref{table:results-small} mirror those obtained on the full datasets, with amplified effects: the proposed loss achieves the best Macro-F1 score on every dataset, and contrastive losses in general perform strongly in this regime.

\begin{figure}[h!]
    \centering
    \includegraphics[width=\textwidth]{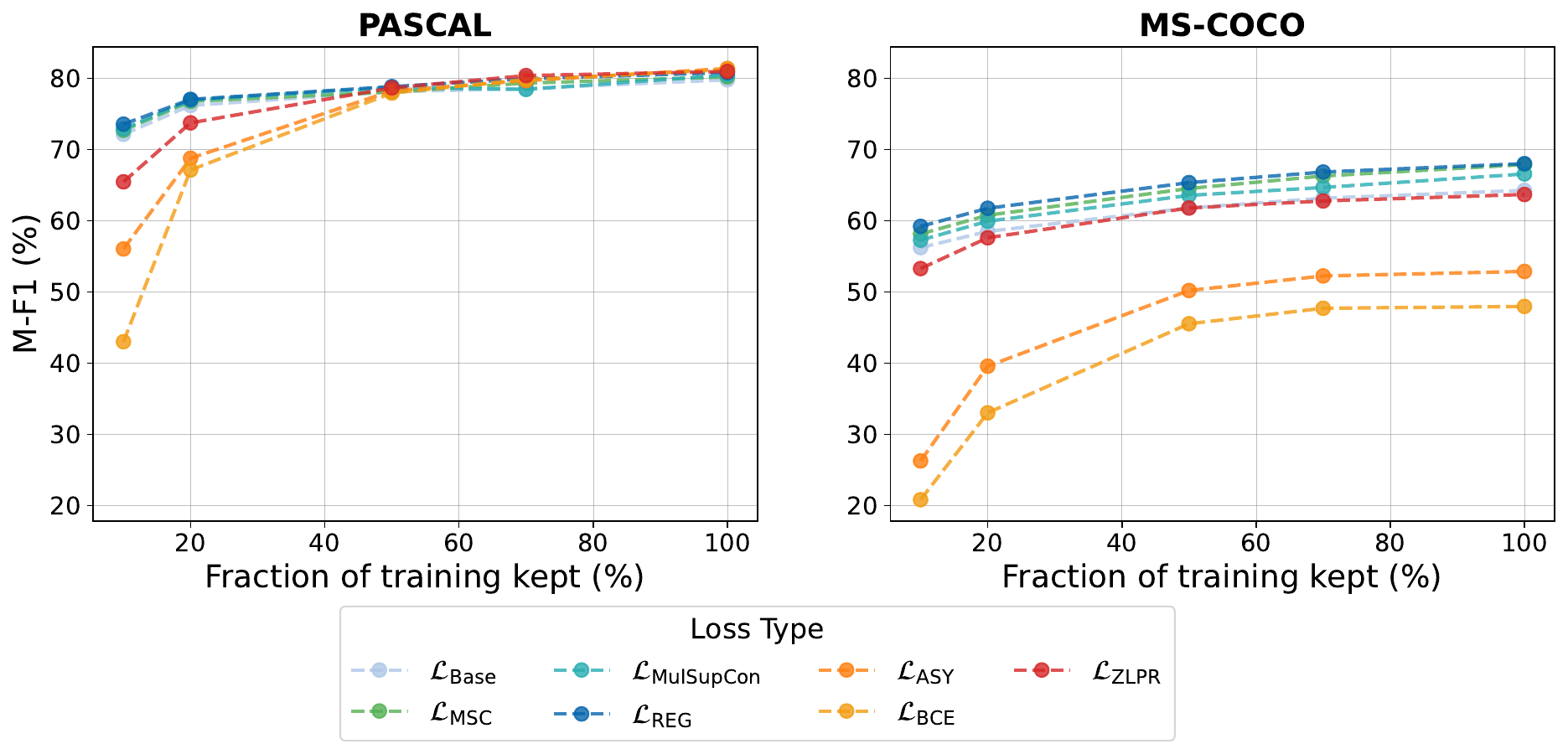}
    \caption{Macro-F1 score as a function of the fraction of training data. Contrastive losses are more robust to data scarcity.}
    \label{fig:H_fraction}
\end{figure}

Fig.~\ref{fig:H_fraction} provides further insight by reporting the Macro-F1 score as a function of the fraction of training data retained on two vision datasets, including \PASCAL{}. Contrastive losses maintain higher Macro-F1 scores as the training fraction decreases. This robustness is particularly visible on \PASCAL{}, whose small size amplifies the effect of reducing the training data. Whereas non-contrastive losses degrade markedly with fewer training examples, contrastive losses better preserve their discriminative power, which makes them a valuable tool for low-data learning, especially in vision tasks.

\subsection{Discussion}
Previous works \cite{exploringcontrastivelearninglongtailed, mulsupcon} showed that accounting for label frequency in positive pairs and adding prototypes improve performance, and our results confirm both findings. Neither $\mathcal{L}_{Base}$ nor $\mathcal{L}_{MulSupCon}$ uses prototypes, and their key difference is that $\mathcal{L}_{MulSupCon}$ accounts for label frequency in positive pairs; $\mathcal{L}_{MulSupCon}$ significantly outperforms $\mathcal{L}_{Base}$, especially in Macro-F1. Likewise, the losses that use prototypes generally achieve higher Macro-F1 scores than those that do not. A further indication is the strong performance on all datasets of our ablated loss $\lwreg$, which combines these insights from previous work.

\subsubsection{Prototypes} The results of $\mathcal{L}_{Proto}$ \cite{proto-only} stand out. A major motivation for supervised contrastive learning is that, unlike standard supervised losses, it intrinsically accounts for label interactions through instance interactions. $\mathcal{L}_{Proto}$ lacks this property, since it only involves interactions between instances and label prototypes. Yet this simple loss, which resembles the cross-entropy loss used in multi-class classification, achieves very high performance. This suggests that the gains of contrastive learning stem not only from modeling label interactions but also from the favorable optimization behavior of contrastive objectives.

\subsubsection{Ablation} To evaluate the impact of the proposed regularization, we compare our full loss $\lreg$ with its ablated version $\lwreg$, which excludes the regularization term. Adding the regularization never degrades Micro-F1 and in some cases improves it, and it improves Macro-F1, particularly on the text datasets.
The regularization can also be applied to the standard \textit{SupCon} loss \cite{supcon}, where it has no significant impact in the multi-class case (see \ref{Appendix:supcon}). We argue that the regularization is useful specifically in the multi-label setting, where the definition of positive pairs is more complex.

\section{Analysis}
\label{sec:analysis}
In this section, we analyze the representations learned by the contrastive losses and the behavior of the proposed regularization term.

\subsection{Representation Analysis}
To analyze the instance representations used for linear evaluation, we rely on two metrics proposed in \cite{metrics}, which contrastive losses have been shown to optimize asymptotically. The alignment loss $\mathcal{L}_{\text{align}}$ in Eq.~\eqref{eq:l-align} is the expected distance between positive pairs; in the multi-label setting, we define positive pairs as instances with exactly the same label set. The uniformity loss $\mathcal{L}_{\text{uniform}}$ in Eq.~\eqref{eq:l-uniform} measures how uniformly the representations are distributed. Since both metrics are sensitive to the norm of the vectors, which can be altered by layer normalization, we $\ell_2$-normalize the representations.
\begin{equation}
    \mathcal{L}_{\text{align}}(f) = \mathbb{E}_{(x,y) \sim p_{\text{pos}}} \left[ \| f(x) - f(y) \|^2_2 \right],
    \label{eq:l-align}
\end{equation}

\begin{equation}
    \mathcal{L}_{\text{uniform}}(f) = \log \mathbb{E}_{(x,y) \sim p_{\text{data}}} \, \left[e^{- 2 \| f(x) - f(y) \|_2^2 }\right],
    \label{eq:l-uniform}
\end{equation}

where \( f \) represents the feature mapping function, \( p_{\text{pos}} \) denotes the distribution of positive pairs, and \( p_{\text{data}} \) denotes the data distribution.

The results in Table~\ref{table:contrastive-sem:1} are consistent with our gradient analysis. On \COCO{} and \PASCAL{}, the proposed loss $\lreg$ obtains the best scores on both $\mathcal{L}_{\text{uniform}}$ and $\mathcal{L}_{\text{align}}$: it improves alignment over the other losses while maintaining competitive uniformity.

\begin{table}[t!]
\centering
\caption{Alignment and uniformity of the representations learned with multi-label contrastive losses on \PASCAL{}, \COCO{}, and \AAPD{}. Lower values are better for both $\mathcal{L}_{\text{uniform}}$ and $\mathcal{L}_{\text{align}}$. The best results are in bold and the second best are underlined.}
\small
\resizebox{.75\textwidth}{!}{ 
\begin{tabular}{lcccccc}
\toprule[1pt] 
\textbf{Loss} & \multicolumn{2}{c}{\textbf{PASCAL}} & \multicolumn{2}{c}{\textbf{COCO}} & \multicolumn{2}{c}{\textbf{AAPD}} \\
\cmidrule(lr){2-3} \cmidrule(lr){4-5} \cmidrule(lr){6-7}
 & $\mathcal{L}_{\text{uniform}}$ & $\mathcal{L}_{\text{align}}$ & $\mathcal{L}_{\text{uniform}}$ & $\mathcal{L}_{\text{align}}$ & $\mathcal{L}_{\text{uniform}}$ & $\mathcal{L}_{\text{align}}$ \\
\midrule
$\mathcal{L}_{\text{Base}}$ & \underline{-1.67} & \underline{0.48} & -1.73 & 0.37 & \underline{-2.86} & 0.72 \\
$\mathcal{L}_{\text{Proto}}$ & -1.61 & 0.54 & -1.61 & 0.38 & -2.07 & \textbf{0.25} \\
$\mathcal{L}_{\text{MulSupCon}}$ & \underline{-1.69} & 0.51 & \underline{-1.77} & \underline{0.35} & \textbf{-3.06} & 0.91 \\
$\mathcal{L}_{\text{MSC}}$ & -1.66 & 0.50 & -1.71 & 0.37 & -2.79 & 0.74 \\
$\lwreg$ & -1.60 & 0.54 & -1.72 & 0.39 & \underline{-2.86} & 0.81 \\
$\lreg$ (Eq. \ref{eq:main_loss})& \textbf{-1.72} & \textbf{0.40} & \textbf{-1.84} & \textbf{0.27} & -2.36 & \underline{0.29} \\
\bottomrule[1pt] 
\end{tabular}}
\vspace{2px}
\label{table:contrastive-sem:1}
\end{table}

On \AAPD{}, compared with its ablated version, $\lreg$ obtains a worse $\mathcal{L}_{\text{uniform}}$ but a better $\mathcal{L}_{\text{align}}$ score, which is consistent with the fact that the regularization term suppresses repulsion between positive pairs. In addition, $\mathcal{L}_{Proto}$ achieves the best $\mathcal{L}_{\text{align}}$ but the worst $\mathcal{L}_{\text{uniform}}$ score, underlining the importance of instance--instance interactions for obtaining uniform representations.

\subsection{Impact of Temperature on the Positive Regularization Ratio}

We now study a quantity specific to the proposed loss, which we call the \textit{positive regularization ratio} (PRR): the fraction of positive pairs for which the regularization term is active. The PRR indicates how often the gradient correction is triggered, and thus whether the regularization stabilizes a substantial share of positive pairs or is only needed occasionally.

We examine how the PRR varies with the temperature of the contrastive loss. Recall that all previous experiments, on both vision and text datasets, use a temperature of 0.1. While this value proved effective in our preliminary evaluations, we leave a thorough study of its impact on performance and representation quality to future work.
The temperature is a crucial hyperparameter of contrastive learning: it scales the similarity scores and thereby determines how strongly pairs are pulled together or pushed apart in the embedding space. We expect lower temperatures, which amplify differences between similarity scores, to yield higher PRR values, and higher temperatures to reduce the PRR by smoothing these differences.

\begin{figure}[t!]
    \centering
    \includegraphics[width=.7\textwidth]{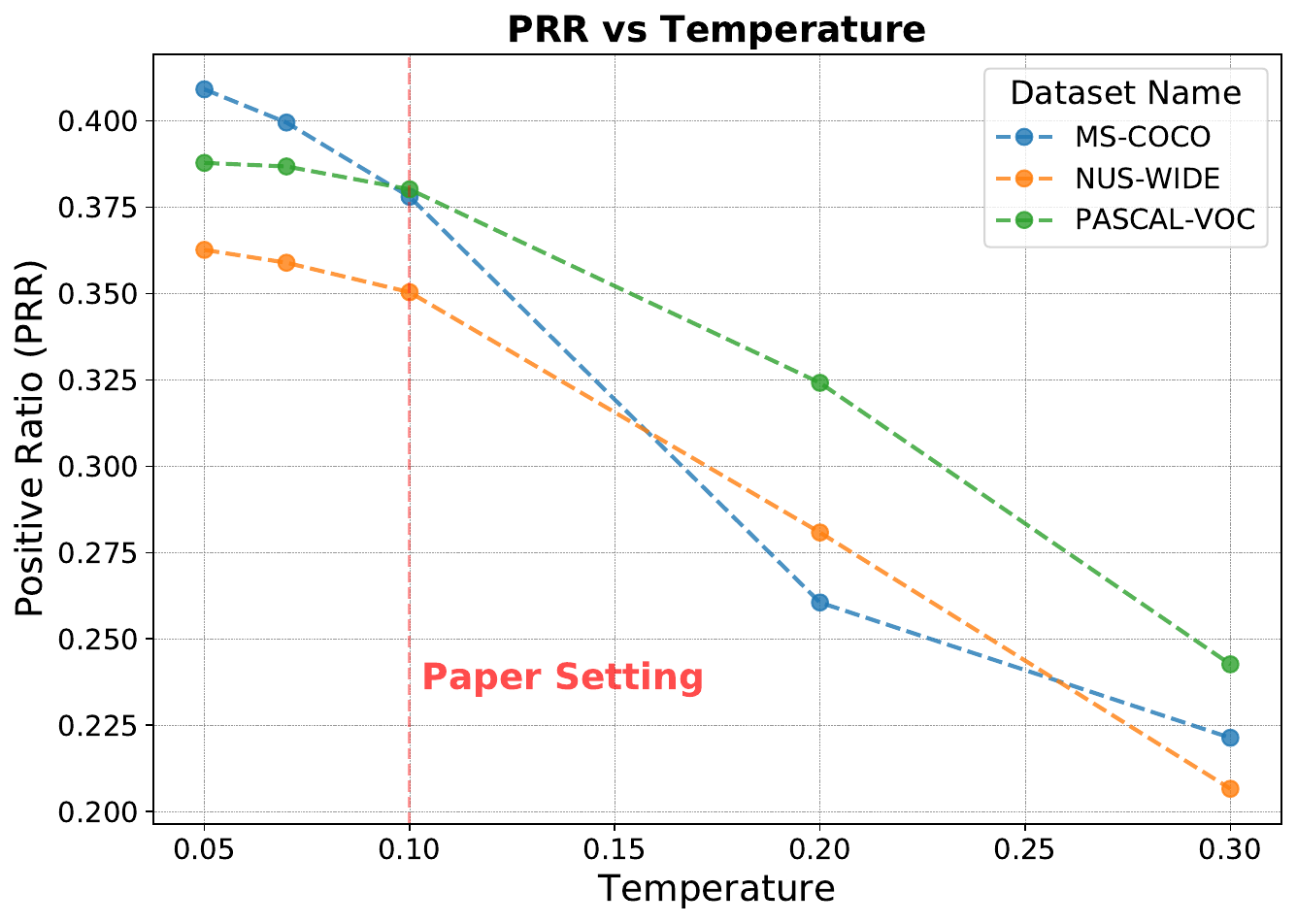}
    \caption{Average positive regularization ratio (PRR) as a function of the temperature of the contrastive loss. Lower temperatures increase the PRR, i.e., the regularization term is activated more often, whereas higher temperatures reduce it by smoothing the differences between similarity scores.}
    \label{tab:temperature-prr}
\end{figure}

Fig.~\ref{tab:temperature-prr} reports the average PRR for several temperature values. The PRR remains substantial across all temperatures; in particular, at the temperature of 0.1 used in this paper, it reaches 38\%, showing that the regularization term is active for a large share of positive pairs and plays an important role in stabilizing their gradients.

\section{Conclusion}
\label{sec:conclusion}

In this paper, we conducted a comprehensive study of contrastive learning for multi-label classification. Our evaluation of contrastive losses across diverse training scenarios, from limited data to full datasets and from vision to text, demonstrates both the effectiveness and the adaptability of these methods.

We showed that the success of contrastive learning in \MLC{} stems not only from modeling label interactions but also from its favorable optimization properties. Our study also revealed important limitations: contrastive approaches are less effective on datasets with few labels and on ranking-based metrics, which calls for targeted advances in these directions.

Building on a gradient analysis, we proposed a new multi-label contrastive loss with a regularization term that prevents positive pairs from repelling each other. This loss integrates key findings from previous studies and is competitive with or outperforms existing losses on \MLC{} benchmarks in both \NLP{} and \CV{}, with the largest gains in Macro-F1 and in low-data regimes.

This work also opens several directions for future research. The strong performance obtained with prototypes alone suggests exploring different weightings of prototype--instance and instance--instance interactions, as well as alternative ways of learning the prototypes. Since the main distinction between MulSupCon and the other losses is that instances with more labels receive a larger weight, weighting instances according to their number of labels or to the number of positive pairs available in the batch is another promising direction.

Finally, new protocols for evaluating the representation space would be valuable. Standard evaluation relies on independent binary classifiers, which account neither for label ranking nor for label interactions; evaluation strategies that reflect these aspects would allow a more faithful assessment of the learned representations.

\section*{CRediT authorship contribution statement}
\textbf{Alexandre Audibert:} Conceptualization, Methodology, Software, Investigation, Visualization, Writing - Original Draft. \textbf{Aur\'elien Gauffre:} Conceptualization, Methodology, Software, Investigation, Visualization, Writing - Original Draft. \textbf{Massih-Reza Amini:} Supervision, Writing - Review \& Editing.

\section*{Declaration of competing interest}
The authors declare that they have no known competing financial interests or personal relationships that could have appeared to influence the work reported in this paper.

\section*{Data availability}
All datasets used in this study are publicly available. The source code is available at: 

{\scriptsize \url{https://github.com/audibert-alexandre-fra/Multi-label-contrastive-comprehensive}}.

\section*{Declaration of generative AI and AI-assisted technologies in the writing process}
During the preparation of this work the authors used \texttt{Mistral} in order to improve the language and readability of the manuscript. After using this tool, the authors reviewed and edited the content as needed and take full responsibility for the content of the publication.

\appendix
\appendix  
\section{Additional Details and Results}

\subsection{Implementation Details}
\label{Appendix:implementationdetails}
The standard losses \(\mathcal{L}_{BCE}\), \(\mathcal{L}_{ASY}\), and \(\mathcal{L}_{ZLPR}\) are trained with the settings listed in Table~\ref{table:hyperparams-standard}. For the contrastive losses, the representation space is evaluated with the averaged outputs of three linear classifiers. For each label, the learning rate and weight decay are selected by grid search on the validation set. 

\begin{table}[h!]
\caption{Hyperparameters of the standard supervised losses.}
\centering
\small
\resizebox{\textwidth}{!}{\begin{tabular}{lcccc}
\toprule[1pt] 
\textbf{Dataset} & \textbf{Epochs} & \textbf{Learning Rate} & \textbf{Batch Size} & \textbf{Optimizer} \\
\cmidrule{1-5}
PASCAL & \{5,10,20\} & 0.003 & 64 & SGD \\
MS-COCO & \{5,10,20\} & 0.01 & 64 & SGD \\
NUS-WIDE & \{5,10,20\} & 0.01 & 64 & SGD \\
AAPD \& AAPD(S) & $\{10, 40, 80, 160 \}$ & $\{5e^{-5}, 2e^{-5} \}$  & $\{32, 64\}$& AdamW \\
RCV1 \& RCV1 (S) & $\{10, 40, 80, 160 \}$ & $\{5e^{-5}, 2e^{-5} \}$  & $\{32, 64\}$& AdamW \\
BGC \& BGC(S) & $\{10, 40, 80, 160 \}$ & $\{5e^{-5}, 2e^{-5} \}$  & $\{32, 64\}$& AdamW \\
\bottomrule[1pt] 
\end{tabular}}%

\vspace{2px}
\label{table:hyperparams-standard}
\end{table}

All hyperparameters for contrastive learning are given in Table~\ref{table:hyperparams-contrastive}.

\begin{table}[h!]
\centering
\caption{Hyperparameters of the supervised contrastive losses.}
\resizebox{\textwidth}{!}{%
\small
\begin{tabular}{lcccccccccc}
\toprule[1pt] 
\textbf{Dataset} & \textbf{Epochs} & \textbf{Learning Rate} & \textbf{Batch Size} & \textbf{$\alpha$} & \textbf{Optimizer} & \textbf{Linear LR} & \textbf{Linear WD} & \textbf{Linear Batch Size} & \textbf{Linear Optimizer} \\
\cmidrule{1-10}
PASCAL & 10 & 0.003 & 64 & 1 & AdamW & $\{1, 1e^{-1}\}$ & $\{1e^{-1},1e^{-2}, 1e^{-4}\}$ & 64 & AdamW \\
MS-COCO & 10 & 0.01 & 64 & 1 & SGD & $\{1, 1e^{-1}\}$ & $\{1e^{-1},1e^{-2}, 1e^{-4}\}$ & 64 & Adam \\
NUS-WIDE & 10 & 0.01 & 64 & 1 &  AdamW & $\{1, 1e^{-1}\}$ & $\{1e^{-1},1e^{-2}, 1e^{-4}\}$ & 64 & RMSprop \\
AAPD & $\{320, 640\}$ & $\{5e^{-5}, 2e^{-5} \}$ & 256 & $\{0, 1\}$ & AdamW & $\{1, 1e^{-1}\}$ & $\{1, 1e^{-1},1e^{-2}, 1e^{-4}, 1e^{-6}\}$ & 256 & AdamW \\
RCV1 & $\{320\}$ & $\{5e^{-5}, 2e^{-5} \}$ & 256 & $\{0, 1\}$& AdamW & $\{1, 1e^{-1}\}$ & $\{1e^{-1},1e^{-2}, 1e^{-4}\}$ & 256 & AdamW \\
BGC & $\{320, 640\}$ & $\{5e^{-5}, 2e^{-5} \}$ & 256 & $\{0, 1\}$& AdamW & $\{1, 1e^{-1}\}$ & $\{1, 1e^{-1},1e^{-2}, 1e^{-4}, 1e^{-6}\}$ & 256 & AdamW \\
AAPD (S) & $\{320, 640\}$ & $\{5e^{-5}, 2e^{-5} \}$ & $\{128, 256\}$ & $\{0, 1\}$& AdamW & $\{1\}$ & $\{1e^{-2}\}$ & 256 & AdamW \\
RCV1 (S) & $\{320, 640\}$ & $\{5e^{-5}, 2e^{-5} \}$ & $\{128, 256\}$ & $\{0, 1\}$& AdamW & $\{1\}$ & $\{1e^{-2}\}$ & 256 & AdamW \\
BGC (S) & $\{320, 640\}$ & $\{5e^{-5}, 2e^{-5} \}$ & $\{128, 256\}$ & $\{0, 1\}$& AdamW & $\{1\}$ & $\{1e^{-2}\}$ & 256 & AdamW \\
\bottomrule[1pt] 
\end{tabular}%
}
\vspace{2px}
\label{table:hyperparams-contrastive}
\end{table}

Regarding $\alpha$, which controls how strongly label overlap affects the re-weighting, no single value performed best across all datasets. We therefore recommend treating $\alpha$ as a tunable hyperparameter and leave a finer study of its influence to future work.

\subsection{Pseudocode}
\label{appendix:pseudocode}
Algorithm~\ref{algo:mscrg} gives the pseudocode of the proposed loss $\lreg$. It proceeds in four steps.
First, it builds a mask excluding self-pairs and computes the pairwise similarity matrix.
Second, it computes the positive-pair weights $\lambda_j^i$ label by label, using a three-dimensional mask of pairs sharing each label, normalized by the anchor's positive count and summed over labels---implementing the label-frequency normalization of Eq.~\eqref{eq:sub_loss}; dividing by the anchor's label count yields the normalized targets $\Lambda_j^i$.
\begin{algorithm}[t!]
    \SetCommentSty{emph}
    \caption{Gradient-Regularized Multi-Label  Learning}
    \KwIn{
      $\mZ \in \mathbb{R}^{n \times d}$: $\ell_2$-normalized output features of the $n$ elements of the batch; \\
      \hspace{11mm} $\mY \in \{0,1\}^{n \times L}$: multi-hot label vectors; \\
      \hspace{11mm} $\tau$: temperature; \\
      \hspace{11mm} $\text{E}$: small constant for numerical stability; \\
    }

    $n \leftarrow \mY\text{.shape[0]}$ \tcp*{\footnotesize batch size}

    \tcc{\footnotesize Mask excluding self-pairs: $\text{mask\_d}[i,j] = 0$ if $i = j$, $1$ otherwise}
    $\text{mask\_d} \leftarrow \text{ones}(n \times n)$\;
    $\text{mask\_d}[0::(n + 1)] \leftarrow 0$ \tcp*{\footnotesize zero on the diagonal}
    $\text{mask\_d} \leftarrow \text{mask\_d}\text{.reshape}(n, n)$\;

    $\text{sim} \leftarrow \mZ \cdot \mZ^T$\;

    \tcc{\footnotesize Positive-pair weights $\lambda_j^i$, computed label by label}
    $\text{mask\_and} \leftarrow \text{einsum}('ac, bc \rightarrow abc', \mY, \mY)$ \tcp*{\small $=1$ iff $a$ and $b$ share label $c$}
    $\text{mask\_and\_d} \leftarrow \text{mask\_d.unsqueeze(2)} \cdot \text{mask\_and}$; 
    $\text{norm} \leftarrow \text{mask\_and\_d}.\text{sum(dim=1)}$; 
    $\lambda \leftarrow \big(\text{mask\_and\_d}/(\text{norm.unsqueeze(1)} + \text{E})\big).\text{sum(dim=1)}$\; 
    $\Lambda \leftarrow \lambda / \mY.\text{sum(dim=1, keepdim=True)}$ \tcp*{\footnotesize normalized weights $\Lambda_b^a$, rows divided by $|\vy_a|$}

    $\text{log}_p, \sigma \leftarrow \text{log\_softmax\_temp}(\text{sim}, \tau, \text{mask\_d})$\;

    \tcc{\footnotesize  Regularization term $\ell_{reg}$: active only for positive pairs with $\sigma_{j,i} > \Lambda_j^i$}
    $\text{mask\_remove\_g} \leftarrow (\Lambda \neq 0)\text{.float()}$ \tcp*{\footnotesize restrict to positive pairs}
    $\text{weight\_reg} \leftarrow \max(-\Lambda + \sigma, 0) \cdot \text{mask\_remove\_g}$ \tcp*{\footnotesize constant coefficient ($\sigma$ is detached)}
    $\text{reg} \leftarrow \text{weight\_reg} \cdot \text{sim}/\tau$ \tcp*{\footnotesize its gradient cancels the repulsive component}

    \KwOut{$-\left(\text{log}_p \cdot \lambda + \text{reg}\right).\text{sum(dim=1)} / n$}

    \label{algo:mscrg}
\end{algorithm}

Third, it takes the normalized similarities $\sigma_{j,i}$ from Algorithm~\ref{algo:logsoftmax}.
Fourth, for positive pairs only ($\Lambda_j^i \neq 0$), it computes the regularization coefficient $\max(0, -\Lambda_j^i + \sigma_{j,i})$; since $\sigma_{j,i}$ is detached, this constant exactly cancels the repulsive gradient for positive pairs whose similarity already exceeds its target (Eqs.~\eqref{eq:derivative-reg-achor} and~\eqref{eq:derivative-reg-positive}).
The loss is finally averaged over the batch.

Algorithm~\ref{algo:logsoftmax} computes a row-wise log-softmax of the similarities scaled by the temperature, in which the denominator only includes the elements selected by the mask, i.e., the set $A(i)$ of all elements of the batch except the anchor. It also returns the corresponding normalized similarities $\sigma_{j,i}$, detached from the computational graph, which are only used to determine when and how strongly the regularization is applied.

\begin{algorithm}[t!]
    \SetCommentSty{emph}
    \caption{LogSoftmax with Temperature and Denominator Mask}
    \KwIn{
      $\text{matrix} \in \mathbb{R}^{n \times n}$: pairwise similarities; \\
      \hspace{11mm} $\tau$: temperature; \\
      \hspace{11mm} $\text{mask} \in \{0,1\}^{n \times n}$: elements included in the denominator (self-pairs excluded); \\
    }

    \tcc{Scale the similarities by the temperature}
    $\text{logits} \leftarrow \text{matrix} / \tau$\;

    \tcc{Exponentiate the scaled similarities}
    $\text{exp}_{\text{logits}} \leftarrow \exp(\text{logits})$\;

    \tcc{Row-wise log-softmax, $\log S(i,j)$, where the denominator sums only over the unmasked elements $k \in A(i)$}
    $\text{log}_p \leftarrow \text{logits} - \log((\text{exp}_{\text{logits}} \cdot \text{mask}).\text{sum(dim=1, keepdim=True)})$\;

    \tcc{Normalized similarities $\sigma_{j,i}$ used to activate the regularization term}
    $\sigma \leftarrow \text{exp}_{\text{logits}} / (\text{exp}_{\text{logits}} \cdot \text{mask}).\text{sum(dim=1, keepdim=True)}$\;

    \tcc{$\sigma$ is detached so that no gradient flows through the regularization coefficient}
    \KwOut{$\text{log}_p$, $\sigma.\text{detach()}$}

    \label{algo:logsoftmax}
\end{algorithm}

\subsection{Performance Across Training Epochs}
\label{appendixB}

We analyze here the behavior of the studied losses as a function of the number of training epochs when the model parameters are randomly initialized. In this setting, it is important to identify the losses whose performance remains stable over longer training, since a loss that quickly overfits would require careful early-stopping strategies and additional validation cost, whereas a loss that improves monotonically with training can be used more reliably in practice.

\begin{figure}[t!]
    \centering
    \includegraphics[width=\textwidth]{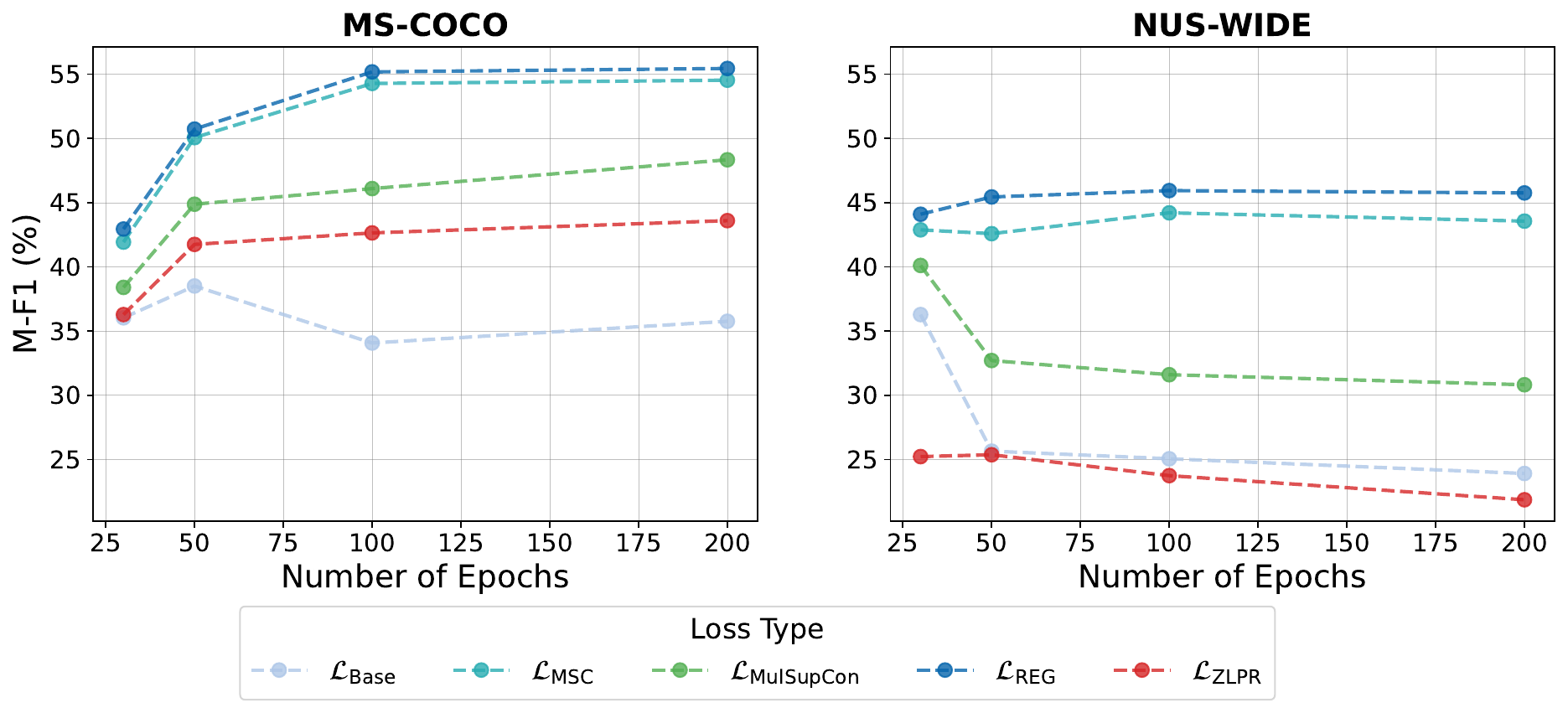}
    \caption{Performance as a function of the number of training epochs with random initialization. Contrastive losses are resistant to overfitting and benefit from longer training, whereas ZLPR shows signs of overfitting, particularly on NUS-WIDE.}
    \label{fig:F_epochs}
\end{figure}

As shown in Fig.~\ref{fig:F_epochs}, contrastive losses are notably resistant to overfitting and consistently benefit from longer training.  This contrasts with the non-contrastive \ZLPR{} loss and the baseline contrastive loss, which show signs of overfitting, particularly on NUS-WIDE: after reaching a peak in the early epochs, their performance steadily decreases as training proceeds, indicating that the models start memorizing the training data rather than learning transferable representations. Interestingly, contrastive losses need only a few epochs to surpass \ZLPR{}.

\subsection{Regularization applied on SupCon Loss}
\label{Appendix:supcon}
The proposed regularization can also be applied to the standard \textit{SupCon} loss \cite{supcon}, which can be written as

\begin{equation}
    \begin{aligned}
    &\mathcal{L}_{\text{SupCon}} = \frac{-1}{|B|} \sum_{\vz_i \in \mathcal{B}} \ell_{\text{SupCon}}(\vz_i) \\
    & \ell_{\text{SupCon}}(\vz_i) = \frac{1}{|P(i)|} \sum_{p \in P(i)} \log \frac{\exp(\vz_i \cdot \vz_p)}{\sum_{a \in A(i)} \exp(\vz_i \cdot \vz_a )}
    \end{aligned}
\end{equation}

Adding the regularization term $\ell_{reg}$ yields the supervised contrastive loss with gradient regularization, denoted by $\mathcal{L}_{SupConReg}$:

\begin{equation}
    \begin{aligned}
    &\mathcal{L}_{\text{SupConReg}} = -\frac{1}{|B|} \sum_{\vz_i \in \mathcal{B}} \hat{\ell}_{\text{SupCon}}(\vz_i) \\
    & \hat{\ell}_{\text{SupCon}}(\vz_i) = \frac{1}{|P(i)|} \sum_{p \in P(i)} \log \frac{\exp(\vz_i \cdot \vz_p)}{\sum_{a \in A(i)} \exp(\vz_i \cdot \vz_a )} + \ell_{reg} (\vz_i)
    \end{aligned}
    \label{eq:supconrg}
\end{equation}

We compare these losses on the 20 Newsgroups dataset \cite{20group}, which contains approximately 18,000 newsgroup posts across 20 topics.

The results in Table~\ref{table:supcon} show that the cross-entropy and \textit{SupCon} losses perform identically, and that adding the regularization brings no noticeable improvement in this multi-class setting.
\begin{table}[h]
\centering
\small
\caption{Comparison of multi-class loss functions on the 20 Newsgroups dataset.}
\begin{tabular}{lccc}
\toprule[1pt] 
\multicolumn{1}{l}{\textbf{Loss}} & $\mathcal{L}_{\text{CE}}$ & $\mathcal{L}_{\text{SupCon}}$ & $\mathcal{L}_{\text{SupConRG}}$ \\
\cmidrule{1-4}
\multicolumn{4}{c}{\textbf{NewsGroup}} \\
\cmidrule{1-4}
Accuracy & 86.60 & 86.68 & 86.64\\
\bottomrule[1pt] 
\end{tabular}%
\vspace{2px}
\label{table:supcon}
\end{table}

\subsection{Ranking Metrics (mAP)}
\label{Appendix:ranking}
We report here the mean average precision (mAP), a common metric in multi-label image classification that measures the ability of a model to rank relevant labels above irrelevant ones.

\begin{table}[b!]
\centering
\caption{Mean average precision (mAP) of multi-label loss functions on PASCAL VOC, MS-COCO, and NUS-WIDE with a pre-trained ResNet-50. The best results, and those not significantly different from them, are in bold.}
\resizebox{.5\textwidth}{!}{ \begin{tabular}{lccc}
\toprule[1pt] 
\multicolumn{1}{l}{} & \multicolumn{1}{c}{VOC} & \multicolumn{1}{c}{COCO} & \multicolumn{1}{c}{NUS-WIDE} \\
\cmidrule(r){2-2} \cmidrule(lr){3-3} \cmidrule(l){4-4}
\cmidrule{1-4}
\multicolumn{4}{c}{\textbf{Non-Contrastive Losses}} \\
\cmidrule{1-4}
$\mathcal{L}_{BCE}$ & 87.44 & 64.18 & 38.00 \\
$\mathcal{L}_{ASY}$ & \textbf{87.54} & 64.77 & 35.37 \\
$\mathcal{L}_{ZLPR}$ & 87.19 & \textbf{71.74} & \textbf{54.85} \\

\cmidrule{1-4}
\multicolumn{4}{c}{\textbf{Contrastive Losses}} \\
\cmidrule{1-4}
$\mathcal{L}_{Base}$ & 80.09 & 65.92 & 51.25 \\
$\mathcal{L}_{Proto}$ & 78.06 & 68.32 & 52.95 \\
$\mathcal{L}_{MulSupCon}$ & 79.58 & 67.69 & 52.47 \\
$\mathcal{L}_{MSC}$ & 80.02 & 68.38 & 52.87 \\
$\lwreg$ & 78.27 & 68.12 & 52.79 \\
$\lreg$ & 79.24 & 68.96 & 53.61 \\

\bottomrule[1pt] 
\end{tabular}}
\vspace{2px}
\label{table:map}
\end{table}

Whereas contrastive losses learn representations based on similarity between instances, \ZLPR{} directly optimizes the relative ordering of labels by penalizing irrelevant labels scored above relevant ones. \ZLPR{} is therefore intrinsically aligned with ranking metrics such as mAP, which is reflected in its superior mAP scores in Table~\ref{table:map}.

\subsection{Results on a Text Dataset with Few Labels}
\label{Appendix:semeval}
To complement the results on PASCAL VOC \cite{pascal}, the vision dataset with the fewest labels, we evaluate the losses on the SemEval-2018 dataset \cite{semeval}, a multi-label emotion classification dataset with only 11 labels (Table~\ref{semeval:dataset}). The results in Table~\ref{table:contrastive-sem} are consistent with those obtained on PASCAL VOC: contrastive losses perform significantly worse, particularly in Micro-F1, except for the proposed loss, which performs on par with the standard losses. On this dataset, the asymmetric loss achieves the best Micro-F1.

\begin{table}[h!]
  \begin{minipage}[t]{0.48\columnwidth}
    \centering
\caption{Statistics of the SemEval-2018 dataset: number of examples (in thousands) in the training, validation, and test sets, number of labels $L$, and average number of labels per example $\overline{L}$.}
    \label{semeval:dataset}
    \setlength{\tabcolsep}{3pt}
  \vspace{2mm}  \begin{tabular}{lccccc}
      \toprule[1pt]
      Dataset & $|\text{Train}|$ & $|\text{Val}|$ & $|\text{Test}|$ & $L$ & $\overline{L}$ \\
      \midrule
      SemEval & 6.6k & 1.1k & 3.2k & 11 & 2.4 \\
      \bottomrule[1pt]
    \end{tabular}
  \end{minipage}\hfill
  \begin{minipage}[t]{0.48\columnwidth}
    \centering
    \caption{Comparison of multi-label loss functions on SemEval-2018 (Hamming loss $\times 10^3$). Best results, and those not significantly different, in bold.}
    \label{table:contrastive-sem}
    \small
    \begin{tabular}{lccc}
      \toprule[1pt]
      & \multicolumn{3}{c}{SemEval} \\
      \cmidrule(r){2-4}
      \textbf{Loss} & $\mu$-F$_1$ & M-F$_1$ & Ham \\
      \midrule
      \multicolumn{4}{c}{\textbf{Non-Contrastive}} \\
      $\mathcal{L}_{BCE}$ & 71.94 & 55.71 & \textbf{11.75} \\
      $\mathcal{L}_{ASY}$ & \textbf{72.43} & \textbf{57.68} & 11.95 \\
      $\mathcal{L}_{ZLPR}$ & 71.52 & 55.70 & \textbf{11.76} \\
      \midrule
      \multicolumn{4}{c}{\textbf{Contrastive}} \\
      $\mathcal{L}_{Base}$ & 70.91 & 57.60 & 12.68 \\
      $\mathcal{L}_{Proto}$ & 69.99 & 56.08 & 12.89 \\
      $\mathcal{L}_{MulSupCon}$ & 70.83 & 57.06 & 12.67 \\
      $\mathcal{L}_{MSC}$ & 69.98 & 56.19 & 12.98 \\
      $\lwreg$ & 70.50 & 56.90 & 12.69 \\
      $\lreg$ & 71.45 & \textbf{57.71} & 12.60 \\
      \bottomrule[1pt]
    \end{tabular}
  \end{minipage}
\end{table}

\clearpage 
\bibliographystyle{elsarticle-num}
\bibliography{custom}

\end{document}